%% file: main.tex
\documentclass{article}
\usepackage{arxiv}

\usepackage[utf8]{inputenc} % allow utf-8 input
\usepackage[T1]{fontenc}    % use 8-bit T1 fonts
\usepackage{hyperref}       % hyperlinks
\usepackage{url}            % simple URL typesetting
\usepackage{booktabs}       % professional-quality tables
\usepackage{amsfonts}       % blackboard math symbols
\usepackage{amssymb}
\usepackage{amsmath}
\usepackage{nicefrac}       % compact symbols for 1/2, etc.
\usepackage{microtype}      % microtypography
\usepackage{cleveref}       % smart cross-referencing
\usepackage{lipsum}         % Can be removed after putting your text content
\usepackage{graphicx}
\usepackage{doi}
\usepackage{glossaries}
\usepackage{xspace}

\makeglossaries
\loadglsentries{./acronyms.tex}

\title{Sampling Strategies for Static Powergrid Models}

\author{ %	
	\href{https://orcid.org/0000-0002-2018-1078}{\includegraphics[scale=0.06]{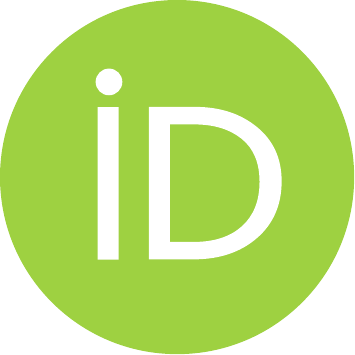}\hspace{1mm}Stephan~Balduin}\\
	OFFIS -- Institute for Information Technology\\
	Escherweg 2, 26121 Oldenburg \\
	\texttt{stephan.balduin@offis.de} \\
\and
	\href{https://orcid.org/0000-0003-2487-7475}{\includegraphics[scale=0.06]{orcid.pdf}\hspace{1mm}Eric~MSP~Veith}\\
	OFFIS -- Institute for Information Technology\\
	Escherweg 2, 26121 Oldenburg \\
	\texttt{eric.veith@offis.de} \\
\and
	\href{https://orcid.org/0000-0003-2340-6807}{\includegraphics[scale=0.06]{orcid.pdf}\hspace{1mm}Sebastian~Lehnhoff}\\
	OFFIS -- Institute for Information Technology\\
	Escherweg 2, 26121 Oldenburg \\
	\texttt{sebastian.lehnhoff@offis.de} \\
}

\renewcommand{\headeright}{Preprint. Under Review}
\renewcommand{\undertitle}{A preprint}
\hypersetup{
pdftitle={Sampling Strategies for Static Powergrid Models},
pdfsubject={Sampling Strategies},
pdfauthor={Stephan~Balduin},
pdfkeywords={machine learning, power grid, powerflow, surrogate models, sampling, correlation},
}

\begin{document}
\maketitle

\begin{abstract}
	\input{chapter/abstract}
\end{abstract}

\keywords{machine learning \and power grid \and powerflow \and surrogate models \and sampling \and correlation}

\input{chapter/01-introduction}
\input{chapter/02-related-work}
\input{chapter/03-doe}
\input{chapter/04-challenges}
\input{chapter/05-correlation-sampling}
\input{chapter/06-evaluation}
\input{chapter/07-conclusion}

\bibliographystyle{unsrt}
\bibliography{main}  

\end{document}

%% file: acronyms.tex
\newacronym{ANN}{ANN}{Artificial Neural Network}
\newacronym{DNN}{DNN}{Deep Neural Network}
\newacronym{DRL}{DRL}{Deep Reinforcement Learning}
\newacronym{PLF}{PLF}{Probabilistic Load Flow}
\newacronym{PF}{PF}{Power Flow}
\newacronym{OPF}{OPF}{Optimal Power Flow}
\newacronym{OPSD}{OPDS}{Open Power System Data}
\newacronym{LV}{LV}{Low Voltage}
\newacronym{MV}{MV}{Medium Voltage}
\newacronym{ML}{ML}{Machine Learning}
\newacronym{SRS}{SRS}{Simple Random Sampling}
\newacronym{LHS}{LHS}{Latin Hypercube Sampling}
\newacronym{DOE}{DOE}{Design of Experiments}
\newacronym{MCS}{MCS}{Monte Carlo Sampling}
\newacronym{PCM}{PCM}{Partial Correlation Matrix}

\newcommand{\ie}{i.\,e.\xspace}
\newcommand{\eg}{e.\,g.\xspace}

%% file: chapter/abstract.tex
Machine learning and computational intelligence technologies gain more and more popularity as possible solution for issues related to the power grid.
One of these issues, the power flow calculation, is an iterative method to compute the voltage magnitudes of the power grid's buses from power values.
Machine learning and, especially, artificial neural networks were successfully used as surrogates for the power flow calculation. 
Artificial neural networks highly rely on the quality and size of the training data, but this aspect of the process is apparently often neglected in the works we found.
However, since the availability of high quality historical data for power grids is limited, we propose the Correlation Sampling algorithm.
We show that this approach is able to cover a larger area of the sampling space compared to different random sampling algorithms from the literature and a copula-based approach, while at the same time inter-dependencies of the inputs are taken into account, which, from the other algorithms, only the copula-based approach does.

%% file: chapter/01-introduction.tex
\section{Introduction}
\label{sec:introduction}

The knowledge about the current state of the power grid is, usually, limited to information about the power generation or consumption of the participants of the grid, either through prognosis or by estimations via default load profiles.
However, a stable grid operation requires a certain frequency level (50 Hz in Europe) and certain voltage levels.
Since only the power values are known, voltage information needs to be calculated, which is done with \gls{PF} analysis.
The \gls{PF} analysis is performed many times during the operation of power grids and the results can be used, \eg, for market analysis or short-term operational planning.
Since the \gls{PF} analysis often requires to perform matrix inversion, a task with a high computational burden, there are many approaches to reduce this computation time.
Active research for improvements of the more traditional methods can be found, \eg, in \cite{grisales2020application,montoya2020power,kontis2019power,yuan2019fast}.
On the other side, the advancement and application of \gls{ML} models for energy systems has also increased in the past two decades.
Mosavi et al. \cite{mosavi2019state} reviewed a broad range of such applications, but \gls{PF} analysis is barely mentioned, possibly due to their selection of relevant papers.
%In 
\cite{hasan2020asurvey} gave an overview about works regarding the closely related \gls{OPF} problem.
\gls{OPF}, however, is only a specific use case for \gls{PF}.
Nevertheless, \gls{ML} for \gls{PF} analysis is a field of active research and, especially, \glspl{ANN} are often used with great performance, \eg, in \cite{nilsson2018machine}, \cite{veerasamy2020novel}, and \cite{donon2019neural}.
However, \glspl{ANN} require a large amount of data, especially when the \gls{ANN} becomes a \gls{DNN}.
In our recent works in \cite{balduin2019towards} and \cite{balduin2020evaluating}, we build different \gls{ML} models, including a \gls{DNN}, as a surrogate model for a \gls{LV} power grid to avoid the costly \gls{PF} analysis.
One issue we identified concerns the availability and generation of training data.
Our data set consisted of time series ranging over one year of data with a 15-minute resolution, resulting in about 35k data points for each time series or, translated to a \gls{ML} task, about 35k training samples.
Those samples need to be further split into training and test samples, which leaves even less data for the training.
On the other side, creating new samples is not a trivial task (as we will show later in this paper), therefore, we decided to have a deeper look at the available data sets and sampling algorithms in the literature.
The rest of this paper is structured as follows.
In \autoref{sec:related-work} we present the results of our investigation and discuss relevant literature. 
In \autoref{sec:doe}, we will have a short look at the basics of sampling strategies and discuss the challenges of power grid sampling in \autoref{sec:challenges}.
In \autoref{sec:correlation-sampling}, we present our \emph{Correlation Sampling} approach to overcome those issues.
We compare and discuss our approach in \autoref{sec:evaluation} and conclude our paper in \autoref{sec:conclusion}. 

%% file: chapter/02-related-work.tex
\section{Related Work}
\label{sec:related-work}
The \gls{OPSD} platform \cite{WIESE2019401} provides a hub for different data sets that can be used for electricity system modeling.
In their work, the authors of the \gls{OPSD} criticize that the quality and accessibility of publicly available data sets is often bad, require different files to download, have poor documentation, or are erroneous.
Most of the data sets provided or linked on \gls{OPSD} concern the power generation.
The available load data sets are either highly aggregated hourly or monthly time series or small-scale household data sets (\eg, there is one with eleven households) that are hardly sufficient to model a distribution grid.
Another good overview of data sets, especially for distribution grids, can be found in the wiki of the openmod initiative\footnote{\url{https://wiki.openmod-initiative.org/wiki/Distribution\_network\_datasets}, retrieved on 07 Apr. 2022}.
The Simbench project \cite{spalthoff2019simbench} provides a large data set, ranging over all German voltage-levels, containing time series for loads and generation.
Finally, there are the IEEE test cases, which mostly focus on North-American-style systems.
Besides using publicly available datasets, which may or may not be synthetic, there are also works that explicitly propose methodologies to create synthetic datasets.
Hülk et al. \cite{hulk2017allocation} proposed a methodology that uses annual consumption data to generate a synthetic data set of the German energy system.
This was extended by Amme et al. 
\cite{amme2018ego} with a focus on the \gls{MV} grid level.
%However, the only large-scale datasets.
Likewise, in the research project SmartNord \cite{Mar2015}, a methodology was proposed to generate synthetic household loads that, once aggregated, follow the German default load profile H0.
Most of these data sets, publicly available as well as synthetically generated, have in common that they range over one year with a time resolution of 15 minutes to one hour.
This may be sufficient for power system modeling, but when it comes to the training of an \gls{DNN}, either multiple data sets need to be combined for the training process or there is not much data left for validation of the model.
When neither of the above mentioned data sets fit or the data set is not large enough, sampling may be the solution.
While this originates in the field of \gls{PLF}, where inputs of the \gls{PF} calculation are modeled as random variables \cite{chen2008probabilistic}, sampling is used in other \gls{PF}-related fields as well.
Cai et al. \cite{cai2013probabilistic} use polynomial normal transformation together with \gls{LHS} to build probability distribution models for \gls{PLF}. 
Their models were able to handle correlated inputs and achieved better results on the IEEE 14-bus and 118-bus systems compared to a \gls{SRS} approach. 
Also in the field of \gls{PLF}, Huang et al. \cite{huang2020improved} sampled with \gls{LHS} as well but used D-vine copulas to model the inter-dependencies of wind speed between four wind farms.
They evaluated the approach on a modified IEEE 33-bus system against \gls{SRS}.
Lei et al. \cite{lei2020data} used a Monte-Carlo simulation approach combined with an interior point algorithm to obtain feasible samples for \gls{OPF}. 
They also did a sample pre-classification to group samples that share the same active constraints.
The test cases were carried out on the IEEE 39, 57, and 118-bus systems as well as on a Polish 2383-bus system.
Some works in the field of \gls{OPF} simply use the base load values provided with most power grid models, % 
\eg, the works in \cite{guha2019machine} and \cite{pan2019deepopf} use 10\% and \cite{zamzam2020learning} even 70\% deviation of the base load, although they, at least, did not sample from a uniform distribution.

In \cite{thayer2020deep}, the authors sampled a variation of the overall consumption and individual scaling factors for each load on the IEEE 14-bus system.
Afterwards, loads are summed up and linearly scaled to match the overall consumption.
Their use case was voltage control based on \gls{DRL}.
Quite similar is the work of Diao et al. \cite{diao2019autonomous}. 
However, the authors used the the base load of the IEEE 14-bus system and created a load fluctuation between 80 \% and 120 \% of the base load values.
From this literature research we conclude that there are a couple of data sets available as well as several ways to generate synthetic data sets.
Unfortunately, those data sets comprise not more than one year of data.
Furthermore, we found different approaches to sample the power grid model, mostly one of the IEEE test cases, directly from different research fields.
While in the \gls{PLF} domain, the works consider actual time series of, \eg, wind farms for generation, this not always the case in the domains \gls{OPF} and voltage control with \gls{DRL}. 
In the works we discussed, the base load if often used for sampling.
The resulting sampling data has a high chance to have completely different distributions than realistic (or synthetic) data, which can affect the quality of a prediction model.
To this end, we propose a methodology that takes into account realistic time series and their inter-dependencies while at the same time preserve the flexibility of the sampling procedure.

%% file: chapter/03-doe.tex
\section{Design of Experiments}
\label{sec:doe}
The power grid model we used is a computer-based simulation model. 
As such, the \gls{DOE} literature would recommend space-filling designs \cite{dean1999design,siebertz2010statistische}.
Space-filling designs aim to spread the sample points for each input evenly in the sample space.
This can be achieved with \gls{MCS} (sometimes also called Simple Random Sampling (SRS)), \ie, using the uniform distribution independently for each input.
Given that enough samples were drawn, this approach creates sampling designs that are nearly orthogonal, \ie, the inputs are uncorrelated.
The \gls{LHS} method is able to provide similar properties while at the same time requires usually less samples to to so.
In general, orthogonality and uniformly distributed inputs are desired properties of a sampling design since they can improve the validity of the prediction model created from that design.
However, there are cases where some of the inputs in the original system-under-investigation are correlated and the power grid is a prime example for this.
To be able to build a model that captures this behavior, the sample distributions for those inputs need to be correlated as well.
One solution is to use Copulas, which were first proposed by \cite{sklar1959fonctions}.
Copulas have the capability to handle marginal distributions of random variables and dependencies separately.
That is the reason increasing number of publications that have to deal with dependencies in power system modeling make use of Copulas, \eg, \cite{huang2020improved,cai2014probabilistic}.

%% file: chapter/04-challenges.tex
\section{Challenges of Power Grid Sampling}
\label{sec:challenges}
\begin{figure}[ht]
	\centering
	\includegraphics[width=0.45\textwidth]{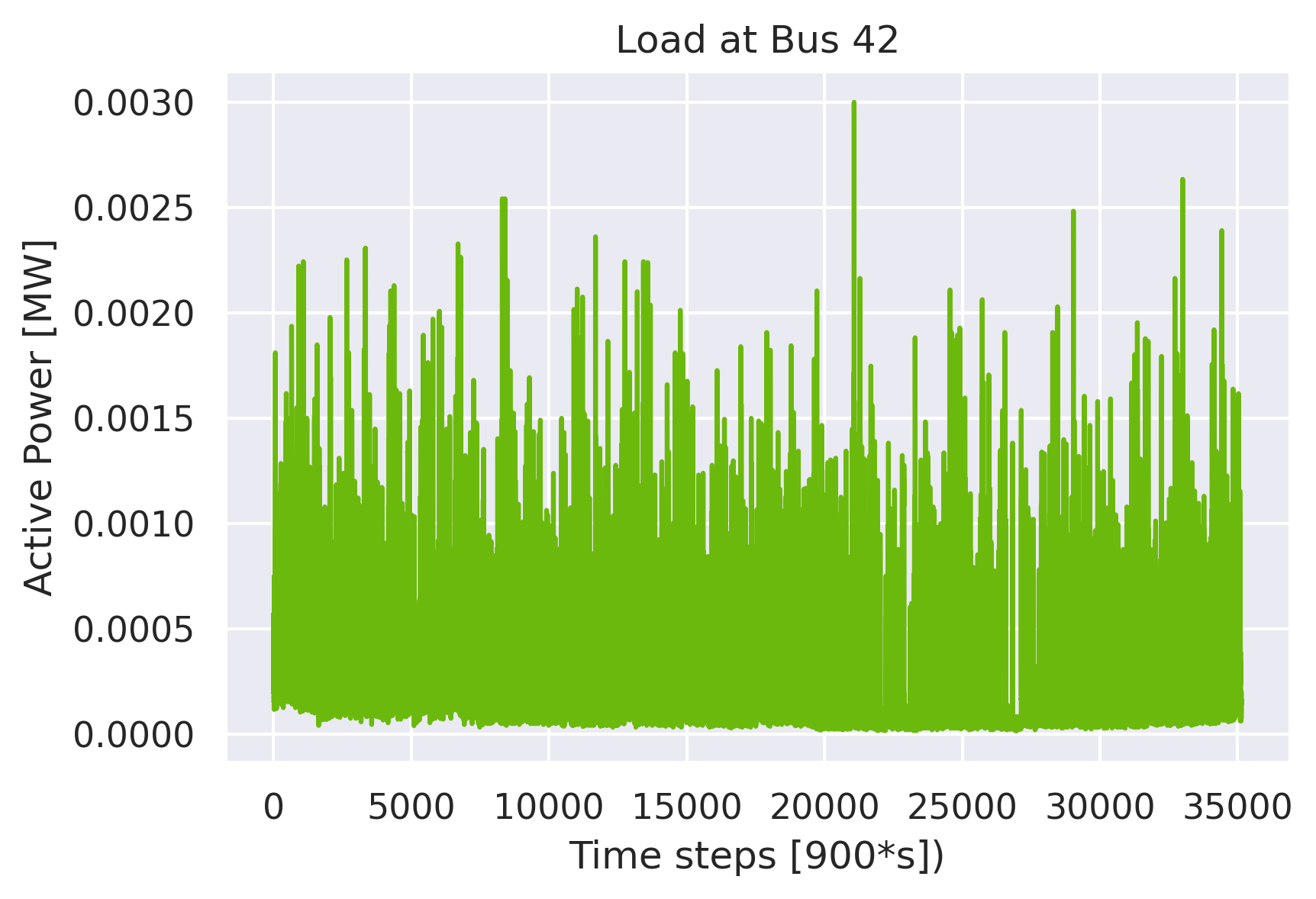}
	\caption{Active power time series of the load connected to bus 42 (randomly selected) over one year of simulated time. Taken from the Simbench grid \texttt{1-LV-rural3--0-sw}.}
	\label{fig:sb-load}
\end{figure}

The power grid is a complex system, \ie, more basic \gls{DOE} approaches for sampling and analysis cannot be applied to the power grid model without modifications.
The complex inter-dependencies between the parts of the power grid make it hard to guarantee properties like orthogonality or uniformly distributed marginal distributions of the inputs without risking a decrease of the quality of the prediction model.
To not consider certain correlations could even lead to ill-conditioned states of the power grid, where the \gls{PF} calculation fails. 
Gerster et al. \cite{gerster2021pointing} investigated sampling strategies for the determination of flexibility potentials at vertical system interconnections.
One of their major conclusions concerns the application uniform sampling for each of the inputs.
With increasing number of inputs, the samples suffer more and more from the convolution problem \cite{gerster2021pointing,bremer2020phase,bremer2018unfolding}, \ie, at the vertical system interconnection point the actually covered space on the P-Q plane gets smaller the more inputs are involved.
Another challenge concerns the definition of the sample space of the inputs.
While for most inputs, zero can be considered as minimum value, the maximum is not clearly defined.
The base value attached to the publicly available power grid models and test cases may serve as reference value.
But it is not a maximum value, since calculating the \gls{PF} using base loads usually results in a healthy or, depending on the test case, slightly violated system state.
It is also not an average value, which can be seen at the distributions of realistic load or generation profiles.
The advantage of using the base load as reference and creating sampling around those values with a certain deviation is that just the grid model itself without any time series data is required.
This makes it convenient if only the general capabilities of an \gls{ML} model should be explored.
But unless the \gls{ML} model is at least evaluated on realistic data, the model may only be a showcase for a certain \gls{ML} algorithm on an environment that happens to be a power grid model.
It does not necessarily imply that this model still performs well if realistic data is used.
To illustrate those issues and to motivate our own approach, we used the aforementioned Simbench project, which provides time series that are at least partially based on real datasets \cite{spalthoff2019simbench}.
In one of our previous works, we already used the Simbench grid with code \texttt{1-LV-rural3--0-sw} and, therefore, we used this grid here as well.
\autoref{fig:sb-load} illustrates the active power of a randomly selected household in the grid.

\begin{figure}[ht]
	\centering
	\includegraphics[width=0.45\textwidth]{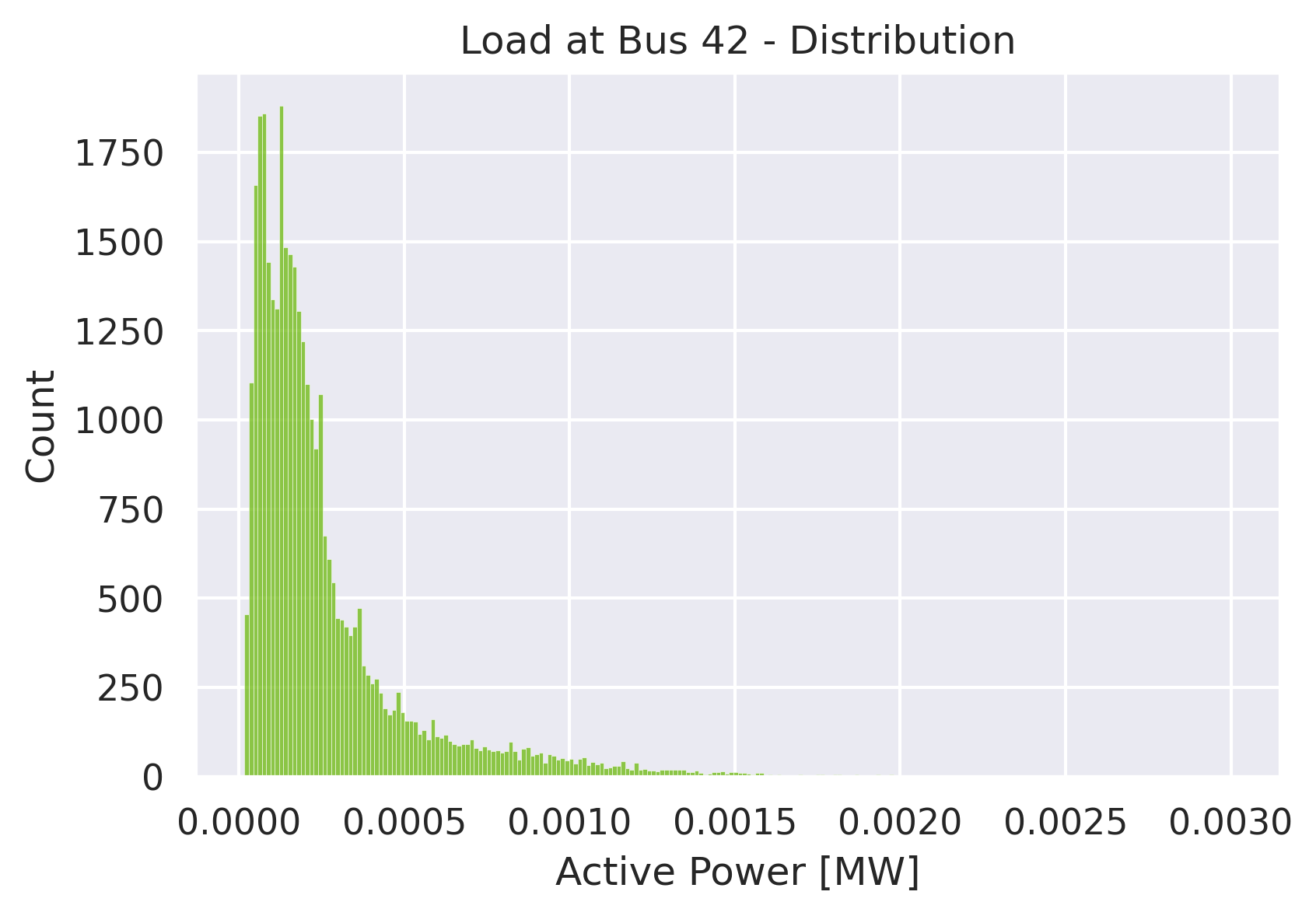}
	\caption{Histogram of the same active power time series as in \autoref{fig:sb-load}.}
	\label{fig:sb-hist}
\end{figure}

The maximum peak power is 3 kW, which is the nominal power of the corresponding load in the grid model.
In \autoref{fig:sb-hist} we plotted the histogram of this time series.
Now it becomes obvious that most of the data is in the range between 0.0 and 0.5 kW.
Actually, the mean value is $\approx$ 0.2657, the standard deviation $\approx$ 0.2781, and the median is at $\approx$ 0.1835.
Sampling around the base load of 3 kW would result in samples, which, although valid, do not represent the original data and, consequently, which do not contain the necessary information for a \gls{ML} model that should make predictions based on realistic input data.

\begin{figure}[ht]
	\centering
	\includegraphics[width=0.45\textwidth]{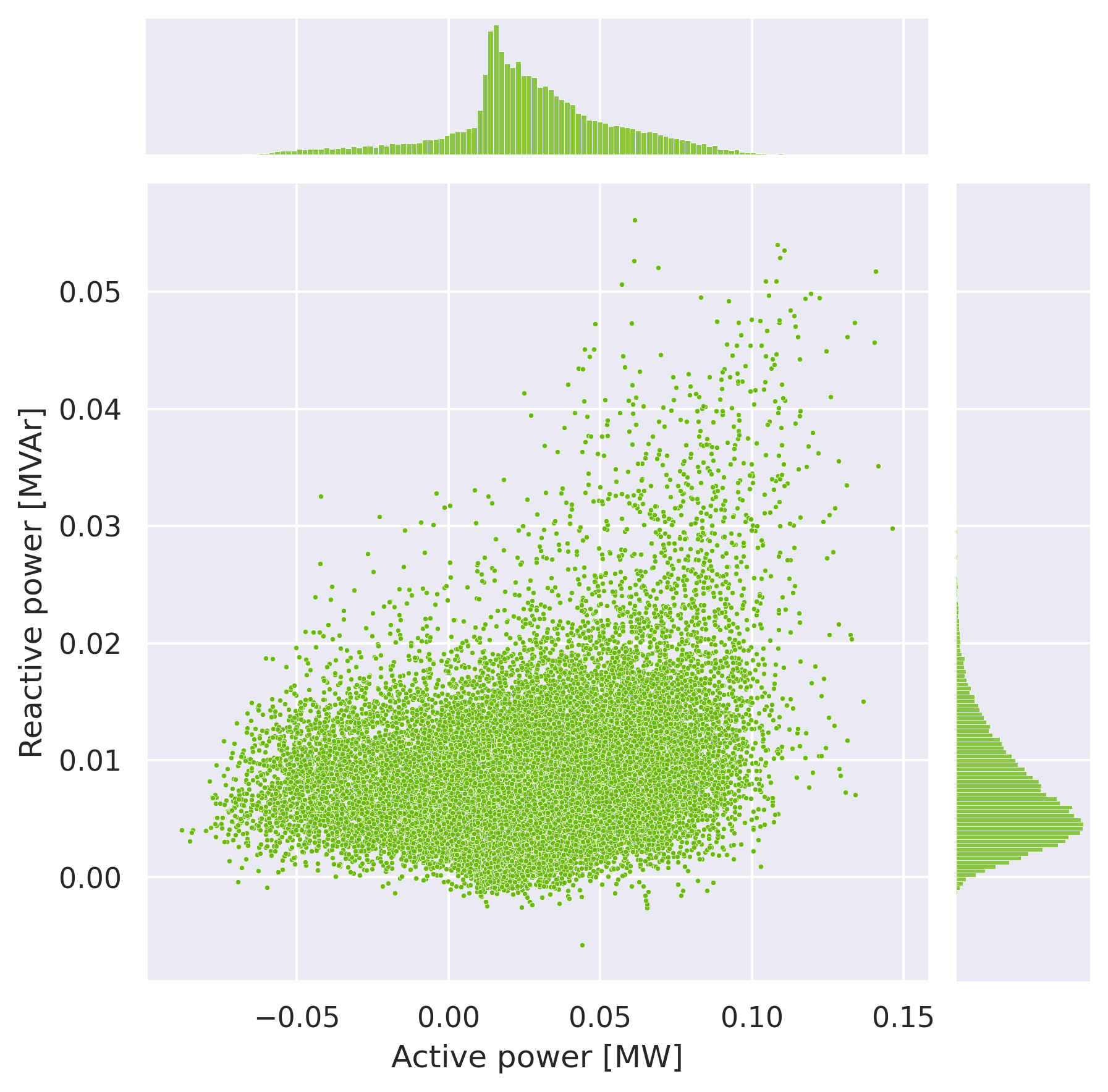}
	\caption{Plot of the active to reactive power plane at the slack bus using the data set that is included in the Simbench grid.}
	\label{fig:sb-pq-dist}
\end{figure}

Next, we simulated the power grid using \emph{pandapower} \cite{pandapower.2018} for one year of simulated time.
We follow \cite{gerster2021comparison} %and looked at the power consumption 
and plotted active against reactive power
at the slack bus %and plotted active power against reactive power 
to get an estimate of the distribution of all of the simulation data and to be able to detect possible convolution problems.
This can be seen in \autoref{fig:sb-pq-dist}.

Now, we wanted to evaluate how well different sampling algorithms from the literature perform for this data set.
We started with two variants of the \gls{SRS} method; the first samples between 0 and the base load (\autoref{eq:uniform}) and the second samples around the base load with a certain $\delta$ (\autoref{eq:srs}).

\begin{equation}
	\label{eq:uniform}
	\mathbf{p}^* \thicksim \mathtt{Uniform} [0, \mathbf{p}_b] 
\end{equation}

\begin{equation}
	\label{eq:srs}	
	\mathbf{p}^* \thicksim \mathtt{Uniform} [(1 - \delta) \cdot \mathbf{p}_b, (1 + \delta)  \cdot \mathbf{p}_b]
\end{equation}

\begin{figure}[ht]
	\centering
	\includegraphics[width=0.45\textwidth]{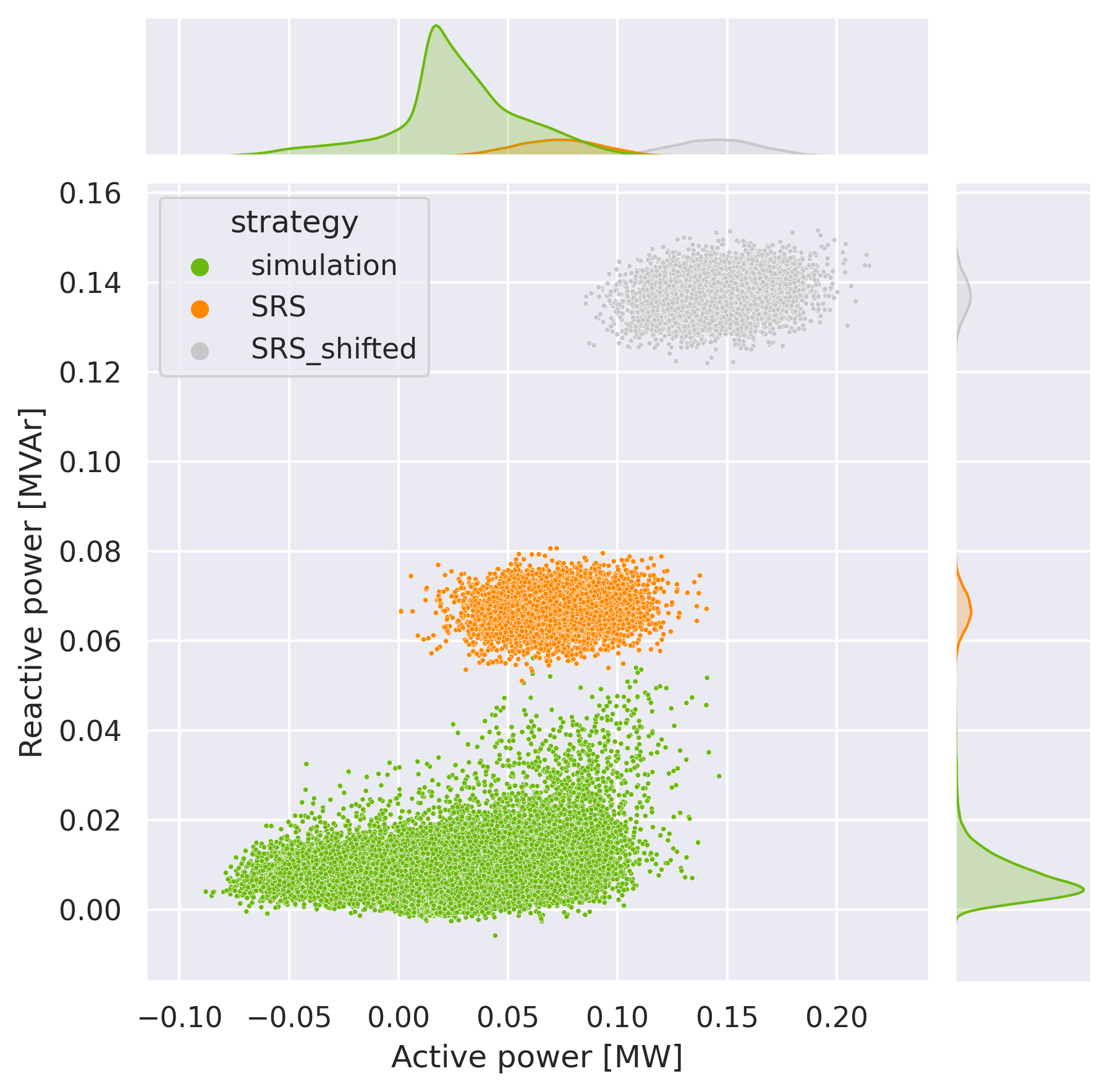}
	\caption{Plot of the active to reactive power plane at the slack bus. The lower dot cloud represents the results from original data sets. The middle dot cloud represents the results from the first \gls{SRS} variant (sampling between zero and the base load) and the upper dot cloud the results from the second \gls{SRS} variant (sampling around the base load).}
	\label{fig:sb-sam-easy}
\end{figure}

Here, $\mathbf{p}_b$ is the vector of base loads in the grid, $\delta$ is the deviation from the base load, and $\mathbf{p}^*$ is the vector of sampled power values.
We used this formulas to generate 5000 samples for active and reactive power consumption as well as active power generation (the generators of the grid in-use had reactive power set to zero) with a $\delta$ of 0.5 in the second case.
Afterwards, we calculated the \gls{PF} for all samples to obtain the active and reactive power for the slack bus, just like above.
The results can be seen in \autoref{fig:sb-sam-easy}.
While all of the samples were feasible (\ie, the \gls{PF} converged), we see that those distributions did not match at all.

A more advanced sampling strategy was used by Thayer et al. \cite{thayer2020deep}.
First, the authors used \autoref{eq:uniform} to sample active power on the interval [0.0, 1.0).
Next, they varied the total active power loading $P'$ uniformly between 60 \% and 140\% of the total active power loading $\overline{P}$ calculated from the base load. 
Each of the loads is scaled linearly with the factor $\overline{P'} / \overline{P^*}$ where $\overline{P^*}$ is the total active power calculated from the samples $\mathbf{p}^*$.
For reactive power $Q$, a power factor $pf$  for each load is drawn uniformly on the interval [0.8, 1.0) and $Q$ is calculated with

\begin{equation}
	Q = P \cdot  \mathtt{tan}(\mathtt{arccos}(pf)) \cdot L, L \in \{-1, 1\}.
\end{equation}

\begin{figure}[ht]
	\centering
	\includegraphics[width=0.45\textwidth]{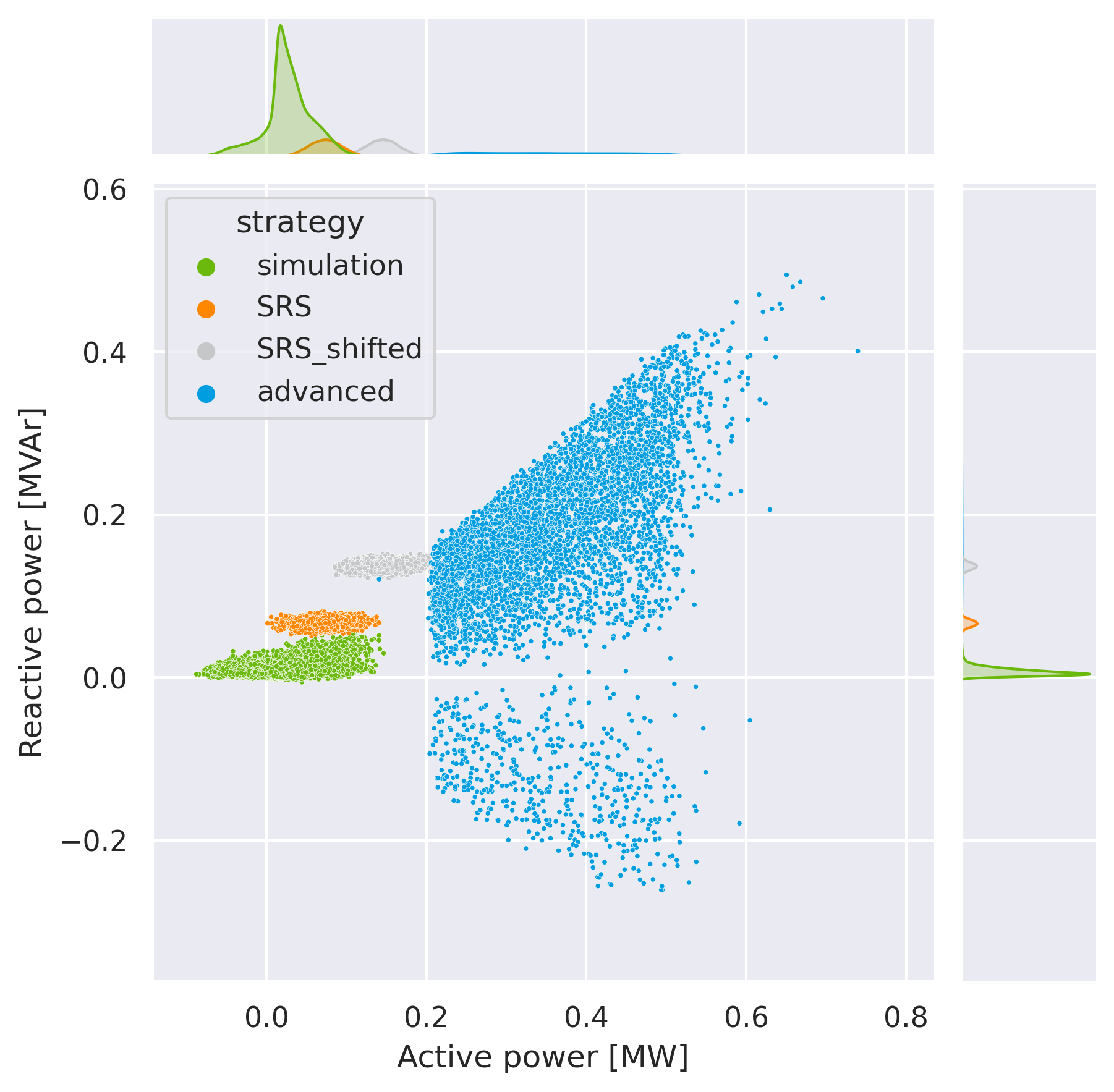}
	\caption{The results of advanced sampling algorithm were added to active-to-reactive power plot.
		The sign-flip of the algorithm in the partly mirrored dot cloud. Overall, the active power values were too high and, therefore, those samples did not cover the area of the original data.}
	\label{fig:sb-sam-adv}
\end{figure}

The factor $L$ is a random variable with a chance of 10\% to be -1 and, therefore, to flip the sign of $Q$.
Like before, we created 5000 samples and calculated the \gls{PF} results.
The $P$ to $Q$ plot is shown in \autoref{fig:sb-sam-adv}.
While a much larger part of the sampling space is covered, the areas of the original data and the sample data were completely different.

\begin{figure}[ht]
	\centering
	\includegraphics[width=0.45\textwidth]{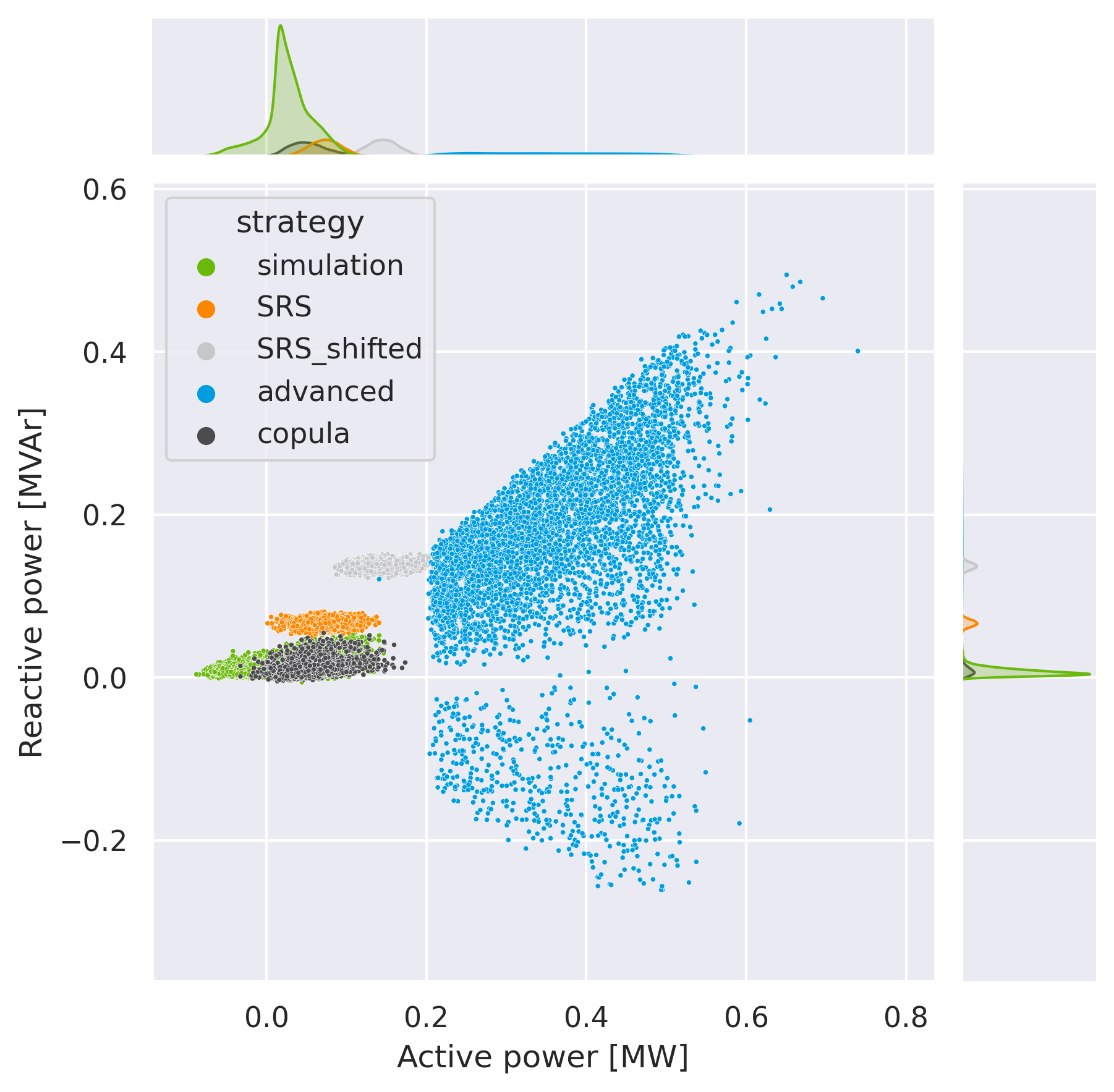}
	\caption{The results of the copula-based sampling were added to the plot.  The darker dot cloud on top of the dot cloud of original data shows that copulas were able to reproduce the behavior of the original data.}
	\label{fig:sb-sam-cop}
\end{figure}

Finally, we created a Gaussian copula to perform the same task. 
We used the python package \emph{copulas}\footnote{\url{https://sdv.dev/}, retrieved on 14 Apr. 2022}, which provides appropriate functions.
The result is shown in \autoref{fig:sb-sam-cop}. 
The copula samples cover most of the space that is covered by the original data as well.

We performed this experiment with different grid models and different time series and got similar results.
Our conclusion is that \gls{SRS}-based approaches are fine when no realistic data sets are available or the model prediction model will not not be used with realistic data sets.
In any other case, copulas allow to create samples that represent the realistic data set.

However, there is one additional concern related to our specific use case.
The copula samples might match the realistic data too well.
One of the desired properties for sampling designs is that the sample points are \emph{evenly spread} over the whole sample space.
Although we don't know the real boundaries of the sampling space, the \gls{SRS}-based approaches cover valid areas of the sampling space that are not covered by the copula samples.
We address this shortcoming with our sampling algorithm.

%% file: chapter/05-correlation-sampling.tex
\section{Correlation Sampling}
\label{sec:correlation-sampling}
The correlation sampling approach consists of two parts.
In the first part, the correlations between the inputs are calculated, which is described in \autoref{sec:correlations}.
Those correlations are used in the second part to create a sampling design, described in \autoref{sec:sampling}

\subsection{Correlations}
\label{sec:correlations}

Naturally, the different entities that are connected to the power grid have inter-dependencies.
Households follow similar patterns although there are different types of profiles.
Photovoltaic modules are heavily dependent on the time of the day and weather conditions like cloudiness and solar radiation, which results in high correlations at spatially close positioned modules.
Correlation can also be found between commercial facilities like different super markets or between several heating devices, which are dependent on temperature conditions.

Utilizing those inter-dependencies is also done in \cite{huang2020improved} to sample wind power plants for \gls{PLF} and in \cite{blank2015reliability} to assess the reliability of coalitions for the provision of ancillary services.
Those inter-dependencies can also be found in the time series data sets for power grids, at least when the data set aims to be realistic.
Therefore, we decided to use correlations, or, more specific partial correlations, to generate samples.
The widely used correlation coefficient by Pearson \cite{benesty2009pearson} is defined as

\begin{equation}
		\mathbf{r}_{XY} = \frac{\mathtt{cov}(X, Y)}{\sigma_X\sigma_Y}
\end{equation}

with $X, Y$ being random variables, $\sigma_X, \sigma_Y$ the standard deviation of $X$ and $Y$ and \texttt{cov} is the covariance.
When more than two random variables are involved, other variables $\mathbf{Z} = (Z_1, \dots, Z_n)$ may have correlation to $X$ and $Y$ as well.
Especially, $Z_i$ might be related to both $X$ and $Y$.
To get the unbiased correlation between $X$ and $Y$, the partial correlation can be calculated.

\begin{equation}
	\mathbf{r}_{XY|Z_i} = \frac{\mathbf{r}_{XY} - \mathbf{r}_{XZ_i} \cdot \mathbf{r}_{YZ_i} } {\sqrt{1 - \mathbf{r}^2_{XZ_i}} \cdot \sqrt{1 - \mathbf{r}^2_{YZ_i}} }
\end{equation}

\begin{figure}[ht]
	\centering
	\includegraphics[width=0.45\textwidth]{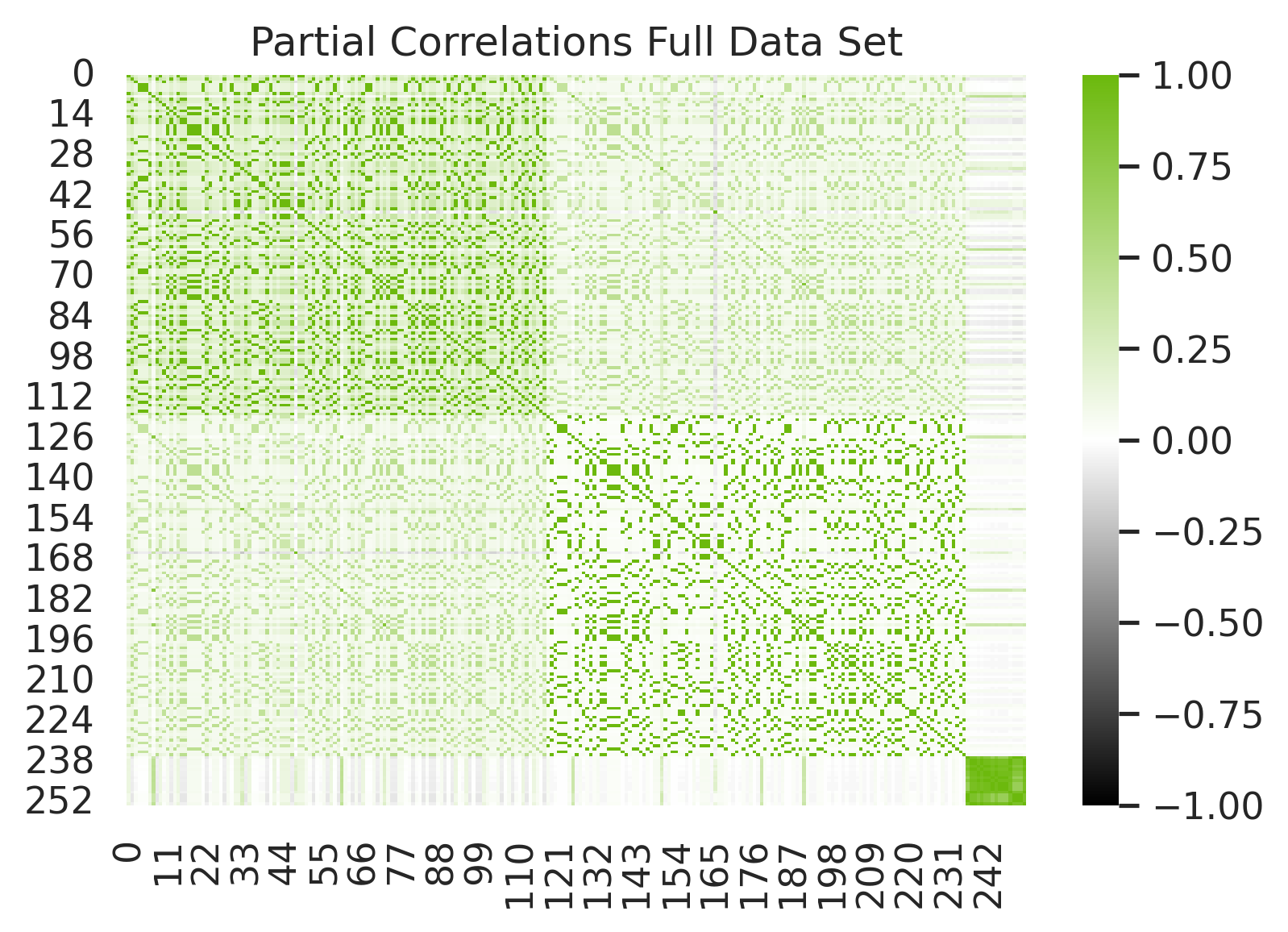}
	\caption{Heat map to illustrate the partial correlations between the inputs of data set provided by the Simbench power grid.}
	\label{fig:sb-full-pc}
\end{figure}

This can be described as two linear regression problems, the first between $Z_i$ and $X$ and the second between $Z_i$ and $Y$ \cite{whittaker2009graphical}.
Since the residuals of those linear regressions are uncorrelated to $Z_i$, the sample correlation can be calculated to obtain the partial correlation between $X$ and $Y$.
We will illustrate this using the data set from the \texttt{1-LV-rural3--0-sw} Simbench grid, which was 
already used in the previous chapter.
The \gls{PCM} $C$ between all of the inputs for the power grid over the full data set, displayed as heat map, can be seen in \autoref{fig:sb-full-pc}.

\begin{figure}[ht]
	\centering
	\includegraphics[width=0.45\textwidth]{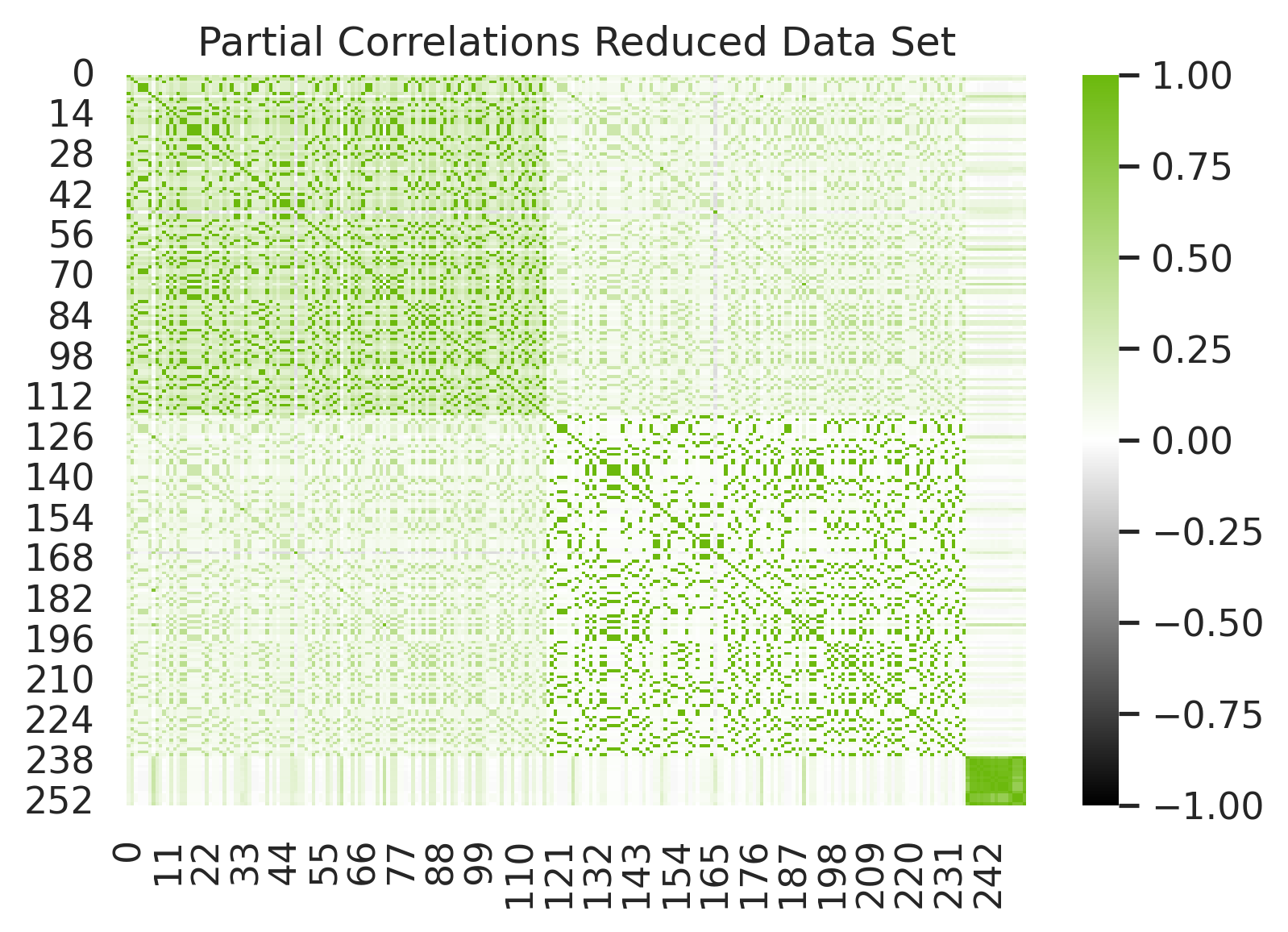}
	\caption{Heat map to illustrate the partial correlations between the inputs. The correlations are calculated from a small subset of the data.}
	\label{fig:sb-reduced-pc}
\end{figure}

Although this \gls{PCM} is sufficient for our sampling algorithm, we still used the whole data set to calculate the correlations.
To overcome this, we selected a subset of the data set containing 2500 samples\footnote{This number is arbitrarily chosen and may only fit the current use case.}  and calculated the \gls{PCM} as $C'$ again.
This heat map can be seen in \autoref{fig:sb-reduced-pc}.

\begin{figure}[ht]
	\centering
	\includegraphics[width=0.4\textwidth]{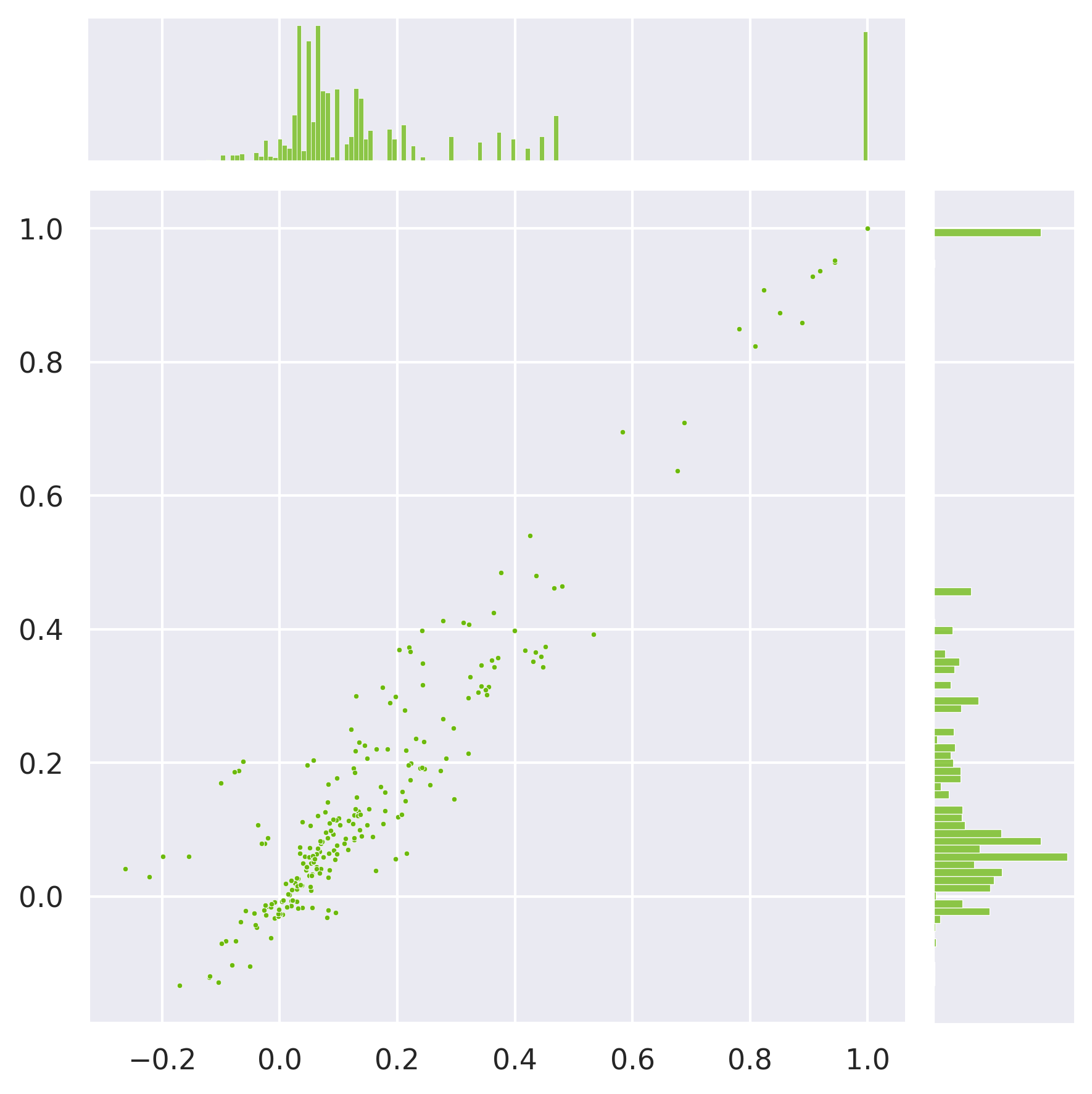}
	\caption{Joint plot over the correlation values from the full PCM and the reduced PCM.}
	\label{fig:sb-joint-full-reduced-pc}
\end{figure}

If you take a close look at both heat maps, you probably recognize similar "patterns".
In fact, those \glspl{PCM} are quite similar with a correlation factor of $\mathbf{r}_{CC'} = 0.98$, with duplicates (the lower left triangle of the matrix) included.
The accumulated point-wise difference $C$ - $C'$ sums up to -391.6 with a mean of -0.006, which indicates that the reduced \gls{PCM} slightly over-estimates positive correlations. 
With a standard deviation of 0.055, we concluded that the reduced \gls{PCM} is similar enough\footnote{This may depend on the data sets in-use. For our use case, the similarity was sufficient}.
\autoref{fig:sb-joint-full-reduced-pc} shows the joint plot between the correlations from $C$ and $C'$.

\subsection{Sampling}
\label{sec:sampling}
The next step concerned how to integrate the \gls{PCM} $C'$ into the sampling procedure.
As it can be seen in \autoref{fig:sb-joint-full-reduced-pc}, most of the partial correlations are lower than 0.5 but there is another cluster between $\approx$ 0.85 and 1.0. 
To not suppress the randomness of the sampling, we defined a threshold $t$ of 0.85 and ignored all correlations that were lower.
Each sample $\mathbf{s}$ is initially generated with the Dirichlet distribution, which was used by Gerster et al. \cite{gerster2021comparison} with good results.
For each sample $\mathbf{s}_i$ in $\mathbf{s}$, all subsequent entries $\mathbf{s}_j$ with $i < j$, are adapted depending on their partial correlation $C'_{ij}$:

\begin{equation}
\label{eq:correlation-sampling}
	\mathbf{s}_j = \begin{cases} 
		\mathbf{s}_j, ~ |C_{ij} | < t \\
		\mathbf{s}_i + \mathbf{s}_j \cdot (1 - C_{ij}), ~ C_{ij} > 0 \\
		1 - \mathbf{s}_i + \mathbf{s}_j \cdot (1 + C_{ij}), ~ C_{ij} < 0
	\end{cases}
\end{equation}

The general idea is to pull the value of sample $\mathbf{s}_j$ towards the value of sample $\mathbf{s}_i$ if they're highly correlated. 
In \autoref{eq:correlation-sampling}, cases two and three account for positive and negative correlation respectively.
We also applied some additional optimizations to better suit the current use case.
First of all, we multiplied $\mathbf{s}_j$ with a normal distributed noise factor of 10\% in cases where the correlation exceeds the threshold to relax the linear dependency towards $\mathbf{s}_i$.
Second, to overcome some of the issues we've seen at the other sampling approaches, we calculated the sum of all values of this sample $\mathbf{s}$ and compared it to an interval [$s_{\mathtt{min}}$, $s_{\mathtt{max}}$].
When $\mathbf{s}$ is not in [$s_{\mathtt{min}}$, $s_{\mathtt{max}}$], $\mathbf{s}$ is discarded and sampled again.
The values $s_{\mathtt{min}}$ and $s_{\mathtt{max}}$ are derived from the data set again, by normalizing each time series individually, then building the sum for each time step, and, finally, assigning the minimum value to $s_{\mathtt{min}}$ and the maximum value to $s_{\mathtt{max}}$.
However, since we used a reduced data set and, therefore, the interval [$s_{\mathtt{min}}$, $s_{\mathtt{max}}$] to small, we extended it by 20 \% in each direction.

%% file: chapter/06-evaluation.tex
\section{Evaluation}
\label{sec:evaluation}

\begin{figure}[ht]
	\centering
	\includegraphics[width=0.45\textwidth]{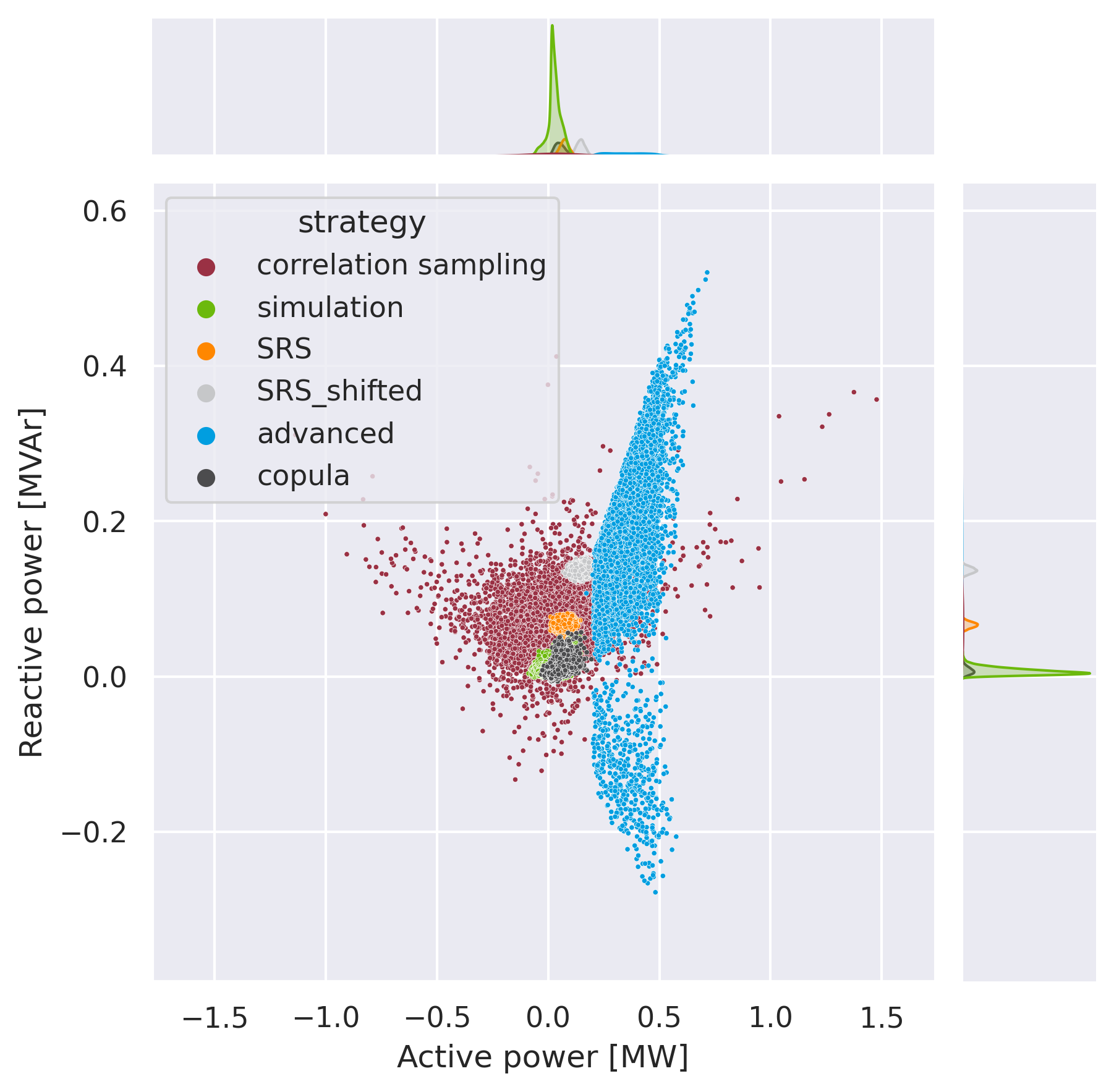}
	\caption{The results of the correlation samples are added to the active-to-reactive power plot. A  large area around the original data is covered, which includes even the areas of most of the other sampling algorithms.}
	\label{fig:sb-sam-cor}
\end{figure}

We used the described methodology to create samples like we did for the other sampling strategies.
The $P$ to $Q$ plot at the slack bus is shown in \autoref{fig:sb-sam-cor}.
It can be seen that the correlation samples not only cover the space of the real data but are also located in the regions largely around the real data.
This even includes most of the space covered by the other sampling strategies.

\begin{figure}[ht]
	\centering
	\includegraphics[width=0.45\textwidth]{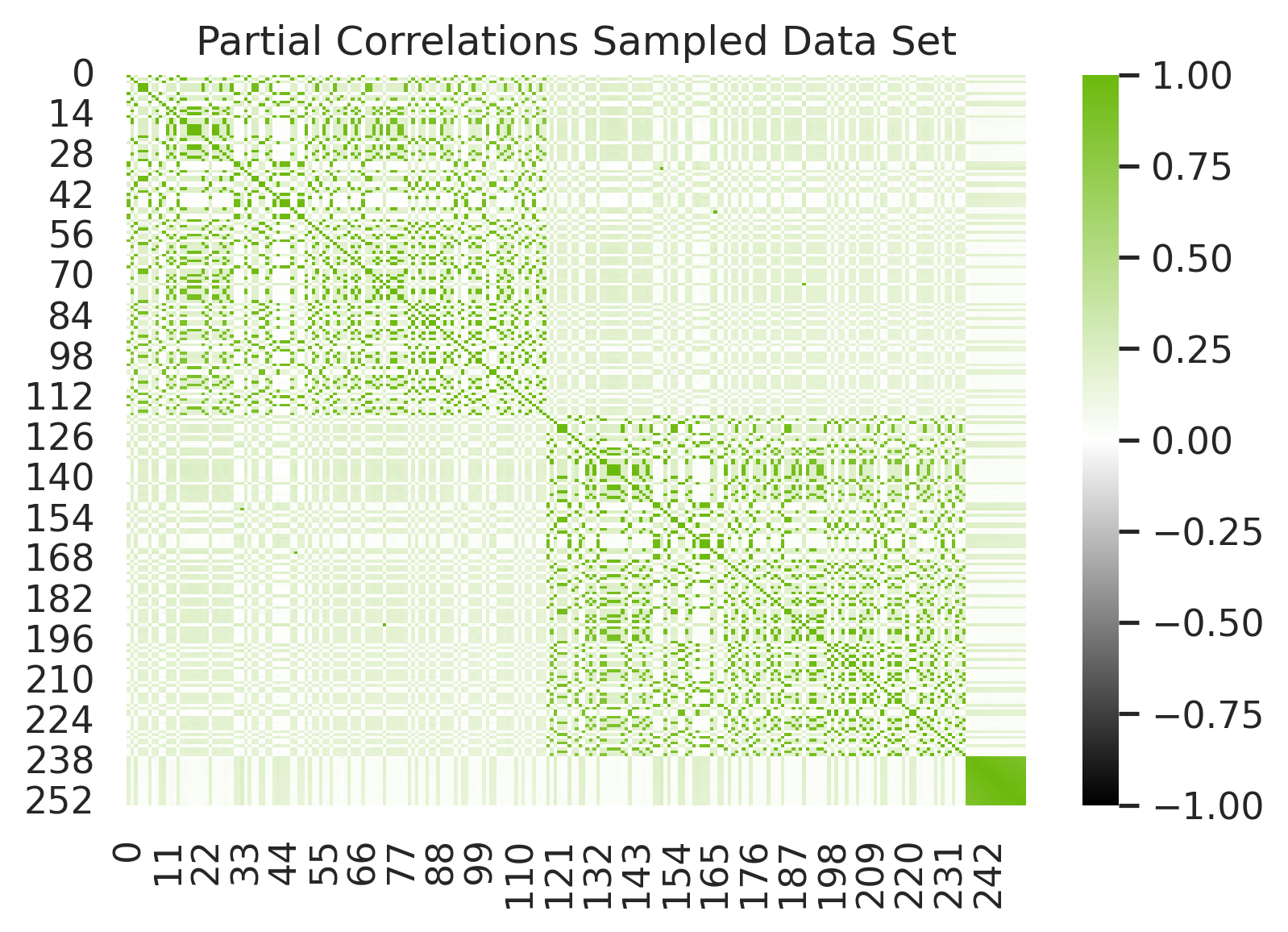}
	\caption{Heat map of the PCMs of the sampling data created with correlation sampling. Visually,  the partial correlation are closer to zero but still follow the same patterns.}
	\label{fig:sbr-pc-sampled-heatmap}
\end{figure}

We also re-calculated the partial correlations of the copula and correlation samples, the latter can be seen in \autoref{fig:sbr-pc-sampled-heatmap}.
The correlation between the copula-sampled \gls{PCM} and the original full-data \gls{PCM} is $\approx$ 0.98, which matches the correlation of the \gls{PCM} of the reduced data set.
For the correlation-sampled \gls{PCM}, the is $\approx$ 0.864, which is less than the reduced data set but still very high. 
However, this difference may be one of the reasons why a larger area is covered. 

\begin{figure}[ht]
	\centering
	\includegraphics[width=0.45\textwidth]{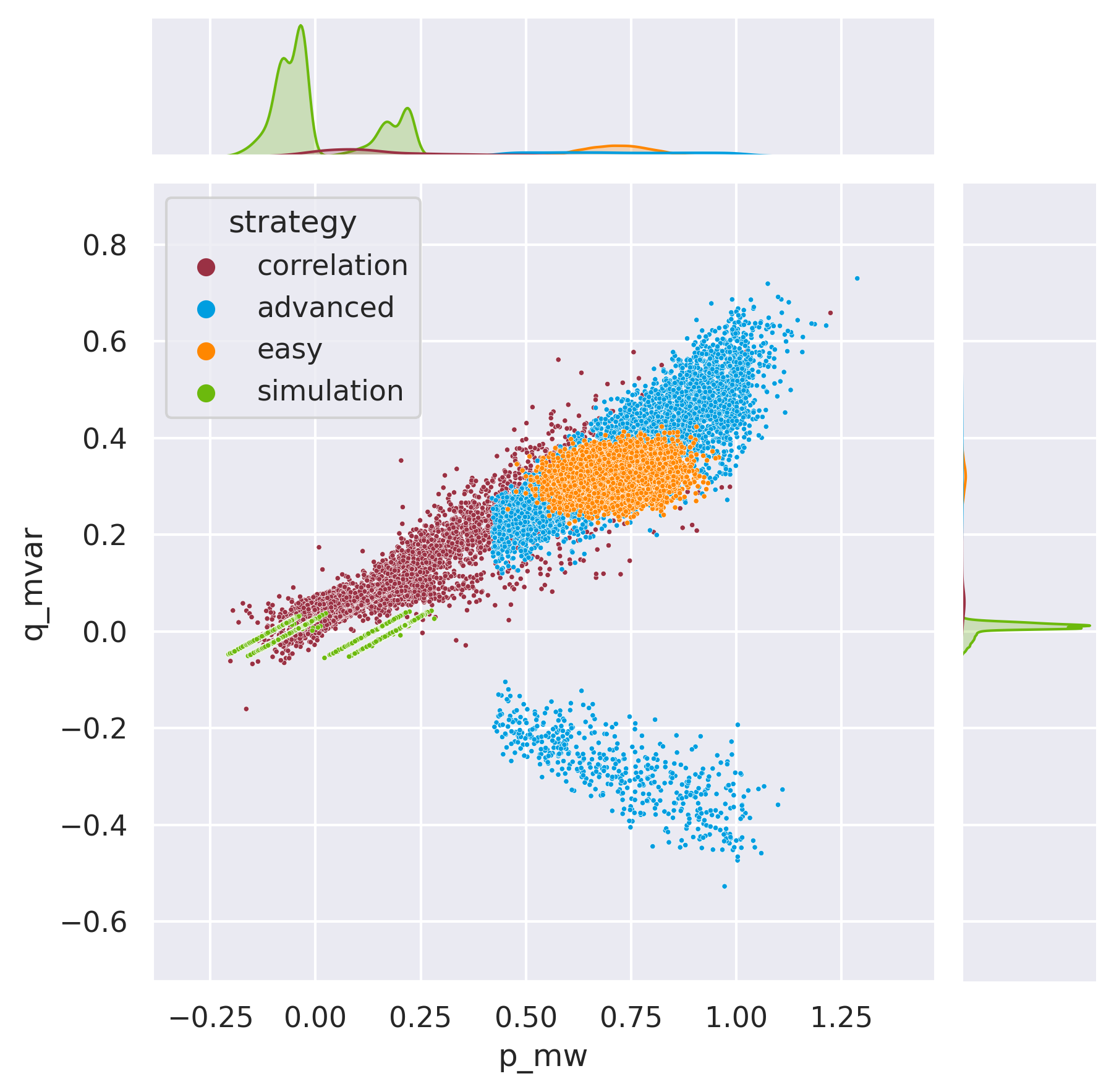}
	\caption{The active-to-reactive power plot with different sampling algorithms for the CIGRE low voltage grid.}
	\label{fig:clv-sam-all}
\end{figure}

We repeated this comparison with another power grid model, the CIGRE low voltage benchmark grid, in combination with synthetic time series data from Smart Nord, since we used this model in our previous works already.
The results are shown in \autoref{fig:clv-sam-all} and this resembles all the issues and conclusions we identified for the Simbench grid, although the data sets are completely independent of each other.

\subsection{Discussion}

The correlation sampling approach is, in its current state, a subject of experimentation.
There are several parameters, like the correlation threshold or the number of samples that are used to calculate the partial correlations, which were determined experimentally and not for a special mathematical reason.
There might be parameters that even lead to better results.
Furthermore, only linear correlations are considered but in the data sets might be nonlinear correlations as well, which could be utilized to get more accurate samples.
On the other side, correlation sampling solved the issue we had with other sampling strategies, as can be seen in the evaluation.
However, this is only a single but important step in the process of our use case: the training of a surrogate model to improve performance in simulation-based analysis.
The evaluation of the correlation sampling approach in this use case is beyond the scope of this paper and will be subject of our next papers.

%% file: chapter/07-conclusion.tex
\section{Conclusion and Future Work}
\label{sec:conclusion}

In this paper, we presented a small literature research about available data sets for power system modeling, where to find them, and discussed some of the issues some of those data sets have.
We also reviewed algorithms from the literature that were used to sample data sets for power grid simulation models and pointed out advantages and disadvantages.
We presented our \emph{Correlation Sampling} approach that aims to combine the advantages of the reviewed sampling algorithms.
Correlation sampling achieved a larger coverage of the sampling space while at the same time taking inter-dependencies between the inputs into account.
Although those first results looks promising, we see a lot of potential for improvements of the algorithm.
Additionally, the data set created with correlation sampling still needs to be used to build a surrogate model, which is the main purpose we developed that algorithm. 
We will present the results from those experiments in a future work.